\documentclass[pmlr]{jmlr}

\usepackage{longtable}
\usepackage{booktabs}

\usepackage{inconsolata}
\usepackage{xcolor} 
\usepackage[framemethod=TikZ]{mdframed}
\usepackage{changepage} 
\definecolor{formalshade}{rgb}{255,255,255}
\mdfdefinestyle{MyFrame}{%
    linewidth=0.5pt,
    linecolor=black,
    backgroundcolor=formalshade,
    innertopmargin=2pt,
    innerleftmargin=2pt,
    innerrightmargin=2pt,
    innerbottommargin=2pt,
    skipabove=5pt
}

\usepackage{subcaption} 
\usepackage{graphicx} 
\usepackage{caption}
\usepackage{multirow}
\usepackage{booktabs}
\usepackage{xcolor}
\usepackage{array}

\makeatletter
\let\Ginclude@graphics\@org@Ginclude@graphics 
\makeatother

\jmlrvolume{260}
\jmlryear{2024}
\jmlrworkshop{ACML 2024}
\jmlrpages{62}
\renewcommand{\thepage}{}

\title[KG-LLM]{Knowledge Graph Large Language Model (KG-LLM) \\ for Link Prediction}

 \author{\Name{Dong Shu}\footnote{Equal contribution} \Email{dongshu2024@u.northwestern.edu}\\
 \addr Northwestern University
 \AND
 \Name{Tianle Chen}\footnotemark[1] \Email{tobychen.razxwasatch@outlook.com}\\
\addr Rutgers University
 \AND
 \Name{Mingyu Jin} \Email{mj939@scarletmail.rutgers.edu}\\
\addr Rutgers University
\AND
\Name{Chong Zhang} \Email{pcczc15@gmail.com}\\
\addr University of Liverpool
\AND
\Name{Mengnan Du}  \Email{mengnan.du@njit.edu} \\
\addr New Jersey Institute of Technology
\AND
\Name{Yongfeng Zhang} \Email{yongfeng.zhang@rutgers.edu} \\
\addr Rutgers University
}

\editors{Vu Nguyen and Hsuan-Tien Lin}

\begin{document}

\maketitle

\begin{abstract}
The task of multi-hop link prediction within knowledge graphs (KGs) stands as a challenge in the field of knowledge graph analysis, as it requires the model to reason through and understand all intermediate connections before making a prediction. In this paper, we introduce the Knowledge Graph Large Language Model (KG-LLM), a novel framework that leverages large language models (LLMs) for knowledge graph tasks. We first convert structured knowledge graph data into natural language and then use these natural language prompts to fine-tune LLMs to enhance multi-hop link prediction in KGs. By converting the KG to natural language prompts, our framework is designed to learn the latent representations of entities and their interrelations. To show the efficacy of the KG-LLM Framework, we fine-tune three leading LLMs within this framework, including Flan-T5, LLaMa2 and Gemma. Further, we explore the framework's potential to provide LLMs with zero-shot capabilities for handling previously unseen prompts. Experimental results show that KG-LLM significantly improves the models’ generalization capabilities, leading to more accurate predictions in unfamiliar scenarios. Our code is available at \url{https://github.com/rutgerswiselab/KG-LLM}.

\end{abstract}


\section{Introduction}

In the domain of data representation and organization, knowledge graphs (KGs) have emerged as a structured and effective methodology, attracting substantial interest in recent years. Although two-node link prediction in KGs has yielded promising results, the multi-hop link prediction remains a difficult task. In real life, multi-hop link prediction plays a crucial role because, often, we are more interested in the relationship between two far-apart entities rather than direct connections. This requires models to reason through intermediate entities and their relationships. A further challenge is the issue of debugging KG model predictions, particularly in the context of discriminative prediction, where the model's lack of explanatory reasoning steps, obscures the origins of errors, diminishing accuracy and performance. Consequently, the development of models capable of generatively and precisely predicting multi-hop links in KGs is a critical challenge.

\begin{figure}
  \centering
  \includegraphics*[width=1\textwidth]{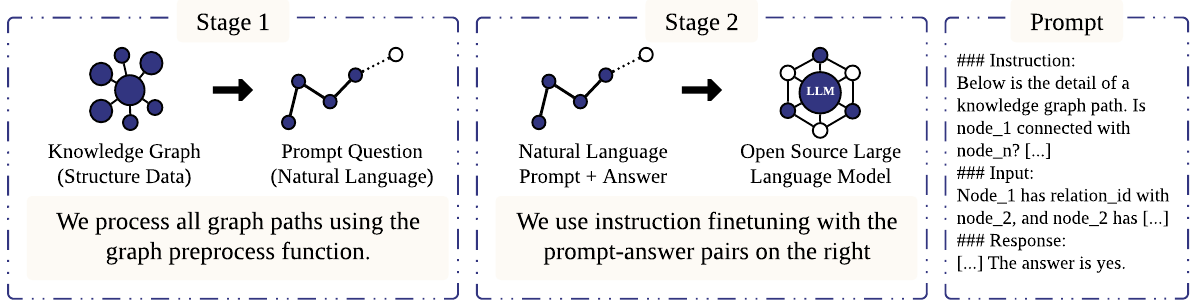}
  \caption{\textbf{A visual overview of KG-LLM framework}}
  \label{KG}
\end{figure}

Historically, approaches to solving tasks related to KGs can trace their origins from embedding-based methods to more recent advancements with LLMs \citep{wang2021survey}. Initially, embedding-based methods played a crucial role, utilizing techniques to represent both entities and relations in a KG as low-dimensional vectors to address the link prediction task by preserving the structural and semantic integrity of the graph \citep{ fan2014transition, lin2015learning}. As the field progressed, the integration of LLMs began to offer new paradigms, leveraging large amounts of data and advanced architectures to further enhance prediction capabilities and semantic understanding in KGs \citep{agarwal2020knowledge, youn2022kglm, yu2022jaket, tang2020bert}. This transition shows a significant improvement from using purely mathematical representations to more context-aware methodologies that better understand the knowledge representation.

Despite these successes, our research highlights three major challenges that prior methodologies have not fully addressed, which our approach aims to solve. First, the predominant focus on discriminative, rather than generative, models and outcomes over reasoning processes underscores a gap in the existing methodologies, highlighting the need for models adept at leveraging reasoning to address multi-hop link prediction in KGs. Secondly, existing approaches predominantly focus on predicting links between two immediate nodes, leaving the field of multi-hop link prediction largely unexplored. This limitation affects the models' ability to navigate and infer connections across extended node sequences. Lastly, the traditional models generally lack generalization abilities, making them less effective when faced with unseen tasks.

To bridge these gaps, we propose the Knowledge Graph Large Language Model (KG-LLM), a novel approach to multi-hop link prediction. As illustrated in Figure \ref{KG}, nodes in KGs are interconnected through specific relations. Initially, our framework takes input from the original KG dataset. After preprocessing, all paths in the KG will transform into chain-of-thought (CoT) prompts \citep{wei2022chain, jin2024impact}, each includes a series of relational statements, which can be represented as \{Node 1 (has relation\_$x$ with) Node 2, Node 2 (has relation\_$y$ with) Node 3, etc.\}. The complexity of the multi-hop problem is determined by its path length and the number of nodes. Through instruction fine-tuning (IFT) \citep{wei2021finetuned} of three Large Language Models (LLMs): Flan-T5 \citep{wei2021finetuned}, LlaMa2 \citep{touvron2023llama}, and Gemma \citep{team2023gemini}, our framework is ready to enhance multi-hop link prediction performance during the testing phase. Moreover, by integrating in-context learning (ICL) \citep{xie2021explanation}, the model not only improves but also has the capacity to tackle unseen prompts, showcasing our method's innovativeness in addressing multi-hop link prediction challenges.

Our study presents the KG-LLM framework as an innovative approach to the multi-hop link prediction task. Our key contributions are:
\begin{itemize}
    \item By converting knowledge graphs into CoT prompts, our framework allows LLMs to better understand and learn the latent representations of entities and their relationships within the knowledge graph.
    \item Our analysis of real-world datasets confirms that our framework improves generative multi-hop link prediction in KGs, underscoring the benefits of incorporating CoT and instruction fine-tuning during training. 

    \item Our findings also indicate that our framework substantially improves the generalizability of LLMs in responding to unseen prompts.
\end{itemize}

\section{Related Work}


Recently, researchers have used Graph Neural Network (GNN) models to solve various graph-related tasks, significantly advancing the field. Among different GNN models, Graph Attention Networks (GATs) have gained attention for their ability to weigh the importance of neighboring nodes, with models like wsGAT \citep{grassia2022wsgat} demonstrating effectiveness in link prediction tasks. Additionally, Graph Convolutional Network (GCN)-based models have shown promising results; ConGLR \citep{lin2022incorporating} leverages context graphs and logical reasoning for improved inductive relation prediction, while ConvRot \citep{le2023knowledge} integrates relational rotation and convolutional techniques to enhance link prediction performance in knowledge graphs (KGs). While the aforementioned approaches have achieved significant success, multi-hop link prediction remains an unsolved challenge.

Other than GNN models, recent development of large language models (LLMs), such as BERT \citep{BERT}, GPT \citep{gpt4}, LLaMA \citep{touvron2023llama}, Gemini \citep{team2023gemini}, and Flan-T5 \citep{wei2021finetuned} has also solved various KGs tasks, including link prediction. The text-to-text training approach makes LLMs particularly suitable for our generative multi-hop link prediction task. Recent studies, concurrent work such as GraphEdit \citep{guo2024graphedit}, MuseGraph \citep{tan2024musegraph}, and InstructGraph \citep{wang2024instructgraph}, have shown that natural language is effective for representing structural data for LLMs. Besides, training on large-scale data makes it possible for LLMs to generalize to unseen tasks or prompts that were not part of its training data \citep{wei2022emergent}. 

Another advantage of LLM-based generative modeling is the Chain-of-Thought (CoT) reasoning ability \citep{weichain}. It provides the flexibility of modifying the instruction, options, and exemplars to allow structured generation and prediction. The Chain-of-Thought reasoning process can be naturally integrated with KGs by translating a reasoning path on a KG into natural language. This flexibility allows us to easily test the model's ability to follow instructions and make decisions based on the provided information. Similarly, In-Context Learning (ICL) \citep{brown2020language} helps LLMs learn from demonstrative examples in the prompt to generate correct answers for the given question. This can also be naturally integrated with KGs. As a result, CoT and ICL enable flexible KG reasoning through natural language.

\begin{figure}
  \centering
  \includegraphics*[width = 1\textwidth]{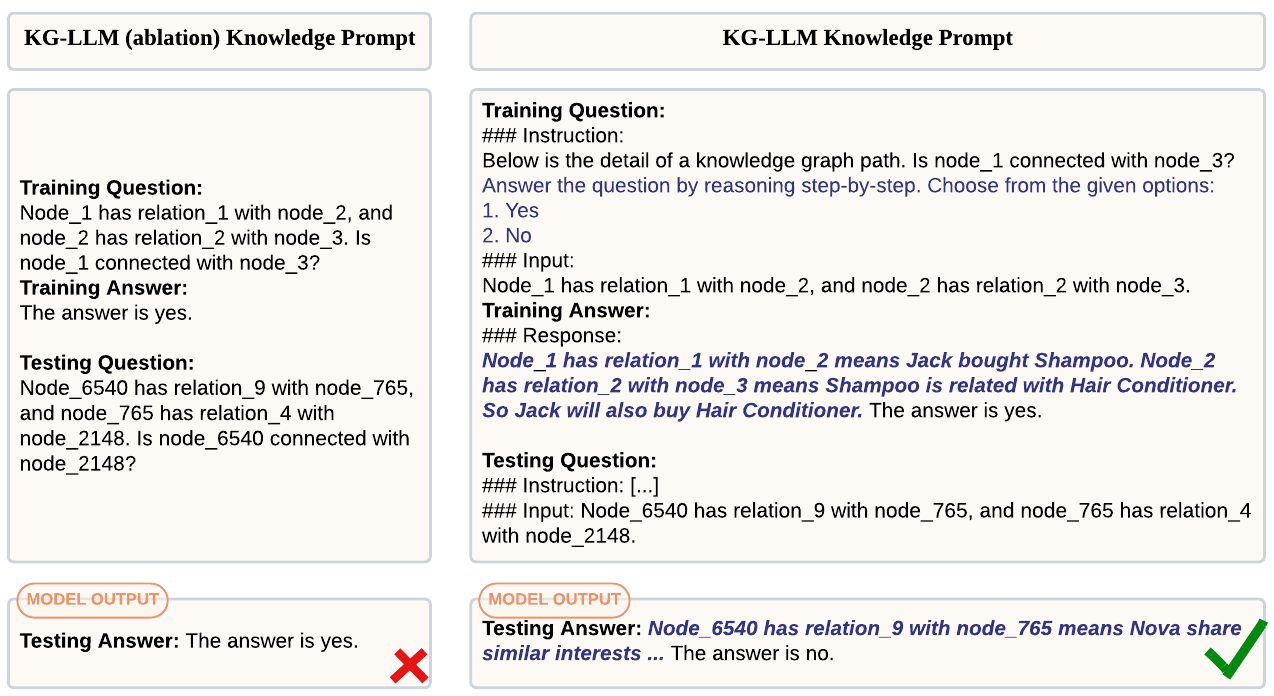}
  \caption{\textbf{An Example of Prompt Used in the Multi-hop Link Prediction Training Process:} Models processed through the ablation framework will be trained using the ablation knowledge prompt (left), whereas models processed via the KG-LLM framework will be trained on the KG-LLM knowledge prompt (right).}
  \label{chain-of-thought}
\end{figure}

\section{Methodology}

In this section, we introduce the proposed KG-LLM framework.

\subsection{Knowledge Graph Definition}
Let $KG = (E, R, L)$ denote a knowledge graph, where $E$ is the set of entities, $R$ is the set of relationships, and $L \subseteq E \times R \times E$ is the set of triples that are edges in the KG. Each triple $(e_i, r, e_{i+1}) \in L$ denotes that there exists a directed edge from entity $e_i$ to entity $e_{i+1}$ via the relationship $r$ \citep{8047276}. 

\subsubsection{Multi-hop Link Prediction}

The task of multi-hop link prediction extends beyond simple link prediction between two nodes. It aims to identify missing connections over multiple relational steps within a knowledge graph. Specifically, given a sequence of observed triples that form a connected path $P_{obs} = {(e_1, r_1, e_2), (e_2, r_2, e_3), \ldots, (e_{n-1}, r_{n-1}, e_n)} \subseteq L$, where each triple $(e_i, r_i, e_{i+1})$ denotes an observed relation $r_i$ between entities $e_i$ and $e_{i+1}$. The objective is to predict whether a missing link $l_{miss} = (e_1, ?, e_n)$ exists by answering True or False \citep{ranganathan2022hoplop, wan2021gaussianpath}.

\subsubsection{Multi-hop Relation Prediction}

The task of multi-hop relation prediction closely aligns with the concept of multi-hop link prediction, with a critical distinction in the question and output. Rather than determining the existence of a missing link $l_{miss}$, this task predicts the relationship. This change shifts the focus from a binary existence query to identifying the specific relationship that binds the entities \citep{nathani2019learning}.

\subsection{Knowledge Prompt}
\label{knowledge_prompt}
\begin{figure}
    \centering
    \includegraphics[width=1\linewidth]{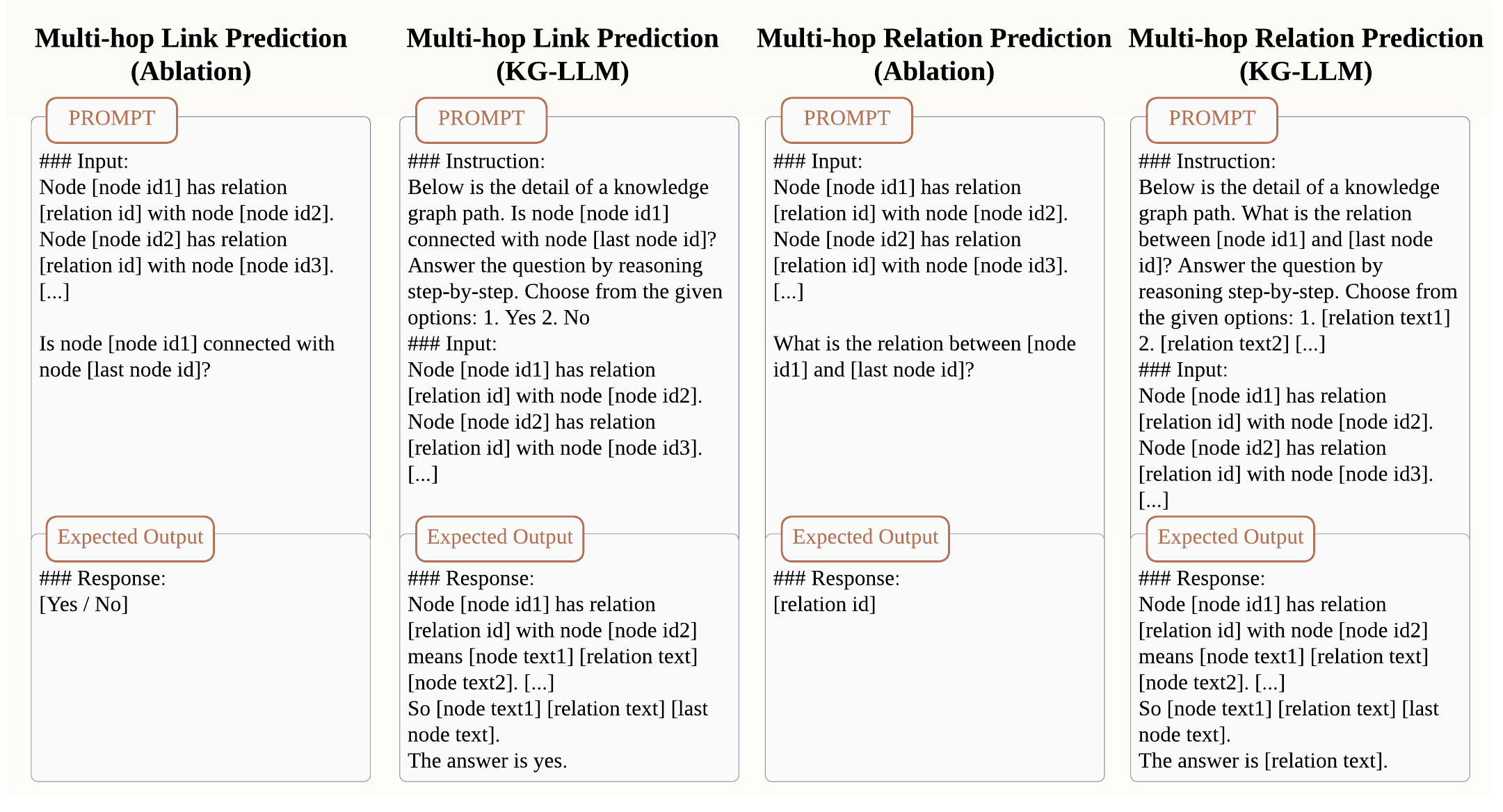}
    \caption{\textbf{Overview of our knowledge prompts in the ablation and KG-LLM Frameworks:} Ablation framework's knowledge prompts are in the first and third columns. KG-LLM framework's knowledge prompts are in the second and fourth columns.}
    \label{fig:prompts}
\end{figure}

The knowledge prompt is a specialized prompt designed for KGs that converts a given sequence of observed triples $P_{obs}$ into natural language. By leveraging the knowledge prompt in the training process, the model can more effectively understand the underlying relationships and patterns present within KGs, thus improving overall performance in multi-hop prediction tasks. In Figure \ref{fig:prompts}, we define the two types of knowledge prompts, KG-LLM knowledge prompt and KG-LLM (ablation) knowledge prompt for both multi-hop link prediction and multi-hop relation prediction.

The two types of prompts demonstrate distinct approaches to enhancing model performance in multi-hop prediction tasks. KG-LLM knowledge prompt adopts a structured format that includes instructions and inputs. This approach involves textualizing node and relation IDs into text based on the dataset and breaking down complex inputs into manageable, concise processes. The KG-LLM instruction falls under the classification category. By listing all possible options in the instructions, LLMs can follow and generate a response based on the given choices. On the other hand, we remove the instruction and textualized IDs in the ablation knowledge prompt and the CoT reasoning process from the expected response. This approach stands out for its clarity and simplicity, providing a good comprehension of the KG and improving prediction accuracy. To illustrate our knowledge prompt better, we provide an example for the multi-hop link prediction task in Figure \ref{chain-of-thought}.


In addition, we adopt the approach of utilizing one-shot ICL learning, specifically tailored to the FLAN-T5-Large, which is our smallest model. This is because, for models of this scale, the impact of utilizing one-shot ICL versus few-shot ICL on accuracy is minimal \citep{brown2020language}. To maintain consistency across our experimental framework, we apply the same one-shot ICL methodology to all LLMs. This uniform approach ensures that our comparative analysis of the models' performances is conducted under equivalent learning conditions. We listed all ICL examples in Appendix \ref{appendix:icl}.

\subsection{KG-LLM Framework}
\label{KG-LLM_Framework}
Our complete KG-LLM Framework is illustrated in Figure \ref{KG}. Initially, the KG is taken as input. Each node is iteratively assigned as the root, and depth-first search (DFS) is used to extract all possible paths. Duplicate paths are then removed, retaining only those with node counts ranging from 2 to 6. This range is based on the ``six degrees of separation'' theory \citep{guare2016six}, which states that any two individuals are, on average, connected through a chain of no more than six intermediaries. The node counts correspond to the number of hops: a single-hop is between two nodes, a two-hop involves three nodes, and so on. These paths are labeled as either positive (there is a connection between the first and last node) and negative (there is no connection) instances. We observed that negative instances outnumbered positive instances, so we randomly reduced the number of negative instances to achieve a balanced dataset. Finally, these paths are converted into KG-LLM and KG-LLM (ablation) knowledge prompts. During the fine-tuning phase, three distinct LLMs are utilized: Flan-T5-Large, LlaMa2-7B, and Gemma-7B. We added all the node IDs and relation IDs as special tokens to the vocabulary of these LLMs. Different fine-tuning techniques are applied for each model within our framework. A global fine-tuning strategy is employed on Flan-T5 to boost its performance. For LlaMa2 and Gemma, a 4-bit quantized LoRA (Low-Rank Adaptation) modification \citep{hu2021lora} is implemented. During the training process, we use the cross-entropy loss function $L$. It calculates the difference between the model's predicted token probabilities and the actual token probabilities in the expected output sequence. In the following equation, $n$ represents the length of the expected output sequence, $x$ stands for the input instruction, and $y_i$ denotes the i-th token in the expected output sequence. 

\begin{equation}
\label{eq:1}
L = -\sum_{i=1}^{n} \log P(y_i | x, y_1, y_2, ..., y_{i-1})
\end{equation}

To evaluate our KG-LLM Frameworks, we train each model twice. As illustrated in Figure \ref{chain-of-thought}, the initial training session employs KG-LLM (ablation) knowledge prompt inputs to establish a baseline. Subsequently, we use instruction finetuning to finetune the original models using KG-LLM knowledge prompt inputs.

After the training phases, we subject each model to two inference task sets, each comprising two sub-tests: non-In-Context Learning (non-ICL) and In-Context Learning (ICL). The primary set of inference tasks is centered around multi-hop link prediction. Conversely, the secondary set probes the models' generalization ability in multi-hop relation prediction, particularly with previously unseen prompts. Through pre- and post-ICL evaluation within each task set, we aim to evaluate the impact of ICL integration across both the KG-LLM (ablation) and KG-LLM frameworks.

\section{Experiments}

In this section, we conduct experiments to evaluate the effectiveness of the proposed KG-LLM frameworks to answer the following several key research questions:

$\bullet$ \textbf{Q1:} Which framework demonstrates superior efficacy in multi-hop link prediction tasks in the absence of ICL?

$\bullet$ \textbf{Q2:} Does incorporating ICL enhance model performance on multi-hop link prediction task?

$\bullet$ \textbf{Q3:} Is the KG-LLM framework capable of equipping models with the ability to navigate unseen prompts during multi-hop relation prediction inferences?

$\bullet$ \textbf{Q4:} Can the application of ICL bolster the models' generalization ability in multi-hop relation prediction tasks?

\subsection{Experimental Setup}

\paragraph{Datasets.}We conduct experiments over $four$ real-world datasets, WN18RR, NELL-995, FB15k-237 and YAGO3-10, which are constructed by the OpenKE library \citep{OpenKE}. All datasets are commonly used for evaluating knowledge graph models in the field of knowledge representation learning. Statistics of the datasets are shown in Table \ref{dataset_table}.

\begin{table}[t]
  \caption{Basic statistics of the experimental datasets.}
  \label{dataset_table}
  \centering
  \begin{tabular}{lrrr}
    \toprule
    Dataset    &\#Entities & \#Triples &\# Relations\\
    \midrule
    WN18RR & 40,943 & 86,835      & 11\\
    NELL-995  & 75,492    & 149,678      & 200 \\
    FB15k-237 & 14,541    & 310,116      & 237 \\
    YAGO3-10  & 123,182   & 1,179,040    & 37 \\
    FB15K     & 14,951    & 592,213      & 1.345 \\ 
    \bottomrule
  \end{tabular}
\end{table}

\paragraph{Task splits.} In the preprocessing stage of each dataset, we randomly selected 80\% of the nodes to construct the training set of KG. Following the steps in section \ref{KG-LLM_Framework}, we constructed the training knowledge prompts. For validation, we randomly split off 20\% of the positive and negative instances from training knowledge prompts. The same procedure was applied to the remaining 20\% of the nodes to create the test set.

\paragraph{Comparing Baselines.} To assess the effectiveness of our KG-LLM frameworks, we compare models' performance across 4 experiments: non-ICL multi-hop link prediction, one-shot ICL multi-hop link prediction, non-ICL multi-hop relation prediction, and one-shot ICL multi-hop relation prediction. In the non-ICL multi-hop link prediction testing, we compare our approach with several traditional methods:
\begin{itemize}
    \item TransE \citep{bordes2013translating} is a traditional distance model that represents relationships as translations in the embedding space.
    \item Analogy \citep{liu2017analogical} can effectively capture knowledge graph structures to improve link prediction.
    \item CompleX \citep{trouillon2016complex} uses complex embeddings to represent both entities and relations, capturing asymmetric relationships.
    \item DistMult \citep{yang2014embedding} represents relations as diagonal matrices for simplicity and efficiency.
    \item RESCAL \citep{nickel2016holographic} uses a tensor factorization method that captures rich interactions between entities and relations.
    \item wsGAT \citep{grassia2022wsgat} is a graph attention network that uses weighted self-attention mechanisms to perform various knowledge graph tasks.
    \item ConGLR \citep{lin2022incorporating} leverages context-aware graph representation learning to enhance link prediction.
    \item ConvRot \citep{le2023knowledge} integrates convolutional networks and rotational embeddings to perform a variety of knowledge graph tasks.
\end{itemize}

\paragraph{Implementation Details.}

We trained each model for 5 epochs on an A40 GPU, and despite limited resources, models still showed promising results. As mentioned in section \ref{KG-LLM_Framework}, we set the maximum complexity of five-hops. We also monitor the input token size to optimize processing efficiency, noting that Flan-T5, with its 512-token capacity, had the smallest token size. Consequently, we tailored our experiments to ensure that the maximum length of input data did not exceed 512 tokens.

\paragraph{Metrics for Multi-hop Link Prediction.}In evaluating the performance of models in multi-hop link prediction tasks, we utilized the Area Under the ROC Curve (AUC) metric \citep{AUC} and the F1 score \citep{F1}. AUC measures the area under the Receiver Operating Characteristic (ROC) curve, which plots the true positive rate against the false positive rate at varying classification thresholds. The threshold is set at a 50\% true positive rate and 50\% false positive rate, as the number of positive and negative data points are equal in the testing case.
A higher AUC value indicates a better ability of the model to differentiate between positive and negative examples. Similarly, the F1 score, ranging from 0 to 1, measures the balance between precision and recall, where higher values represent better performance. For the performance tables presented below, the best performance is indicated in \textbf{bold}, while the second-best performance is indicated with \underline{underline}.

\paragraph{Metrics for Multi-hop Relation Prediction.}
We use accuracy as the performance metric for the multi-hop relation prediction task, which provides an overall measure of the model's correctness, calculated as the percentage of test cases where the true relation is predicted correctly.


\begin{table}[ht!]
\centering
\caption{Multi-hop Link Prediction \textbf{w/o} In-Context Learning}
\setlength{\tabcolsep}{2.5pt}
\begin{tabular}{ccccccccc}
\toprule
Datasets & \multicolumn{2}{c}{\textbf{WN18RR}} & \multicolumn{2}{c}{\textbf{NELL-995}} & \multicolumn{2}{c}{\textbf{FB15k-237}} & \multicolumn{2}{c}{\textbf{YAGO3-10}}\\ 
\cmidrule(lr){1-1}\cmidrule(lr){2-3}\cmidrule(lr){4-5}\cmidrule(lr){6-7}\cmidrule(lr){8-9}
Metrics & F1  & AUC  & F1  & AUC & F1  & AUC & F1  & AUC \\
\cmidrule{1-9}
TransE &0.37 &0.47 &0.26 &0.43 &0.29 & 0.48 & 0.34 & 0.51\\
Analogy &0.61 &0.52 &0.29 &0.47 &0.35 & 0.54 & 0.39 & 0.56\\
CompleX &0.60 &0.51 &0.29 &0.49 &0.32 & 0.51 & 0.36 & 0.53\\
DistMult &0.56 &0.48 &0.25 &0.44 &0.30 & 0.50 & 0.35 & 0.52\\
Rescal &0.61 &0.50 &0.59 &0.53 &0.43 & 0.58 & 0.47 & 0.61\\
wsGAT &0.69 &0.71 &0.62 &0.67 &0.50 & 0.63 & 0.54 & 0.66\\
ConGLR &0.74 &0.69 &0.66 &0.64 &0.63 & 0.68 & 0.69 & 0.61\\
ConvRot &0.75 &0.77 &0.72 &0.66 &0.67 & 0.62 & 0.62 & 0.63\\
\cmidrule{1-9}
Flan-T5 (Ablation) &0.63 &0.67 &0.60 &0.66 &0.63 & 0.67 & 0.68 & 0.70\\
LLaMa2 (Ablation) &0.74 &0.72 &0.71 &0.73 &0.69 & 0.72 & 0.76 & 0.75\\
Gemma (Ablation) &0.76 &0.73 &0.72 &0.71 &0.65 & 0.73 & 0.78 & 0.76\\
\cmidrule{1-9}
Flan-T5 (KG-LLM) & 0.73 & 0.71 & 0.70 & 0.72 & 0.66 & 0.70 & 0.74 & 0.75\\
LLaMa2 (KG-LLM) & \underline{0.82} & \bf0.83 & \underline{0.81} &\underline{0.80} & \underline{0.73} & \underline{0.76} & \textbf{0.85} & \textbf{0.86}\\
Gemma (KG-LLM) & \bf 0.84 & \underline{0.81} &\bf0.82 &\bf0.83 &\textbf{0.79} & \textbf{0.81} & \underline{0.82} & \underline{0.83}\\

\bottomrule
\end{tabular}
\label{link_without}
\end{table}

\begin{figure*}[!ht]
	\centering  
        \captionsetup[subfigure]{skip=2pt}
        
        \subfigure[wsGAT]{\includegraphics[width=0.32\linewidth]{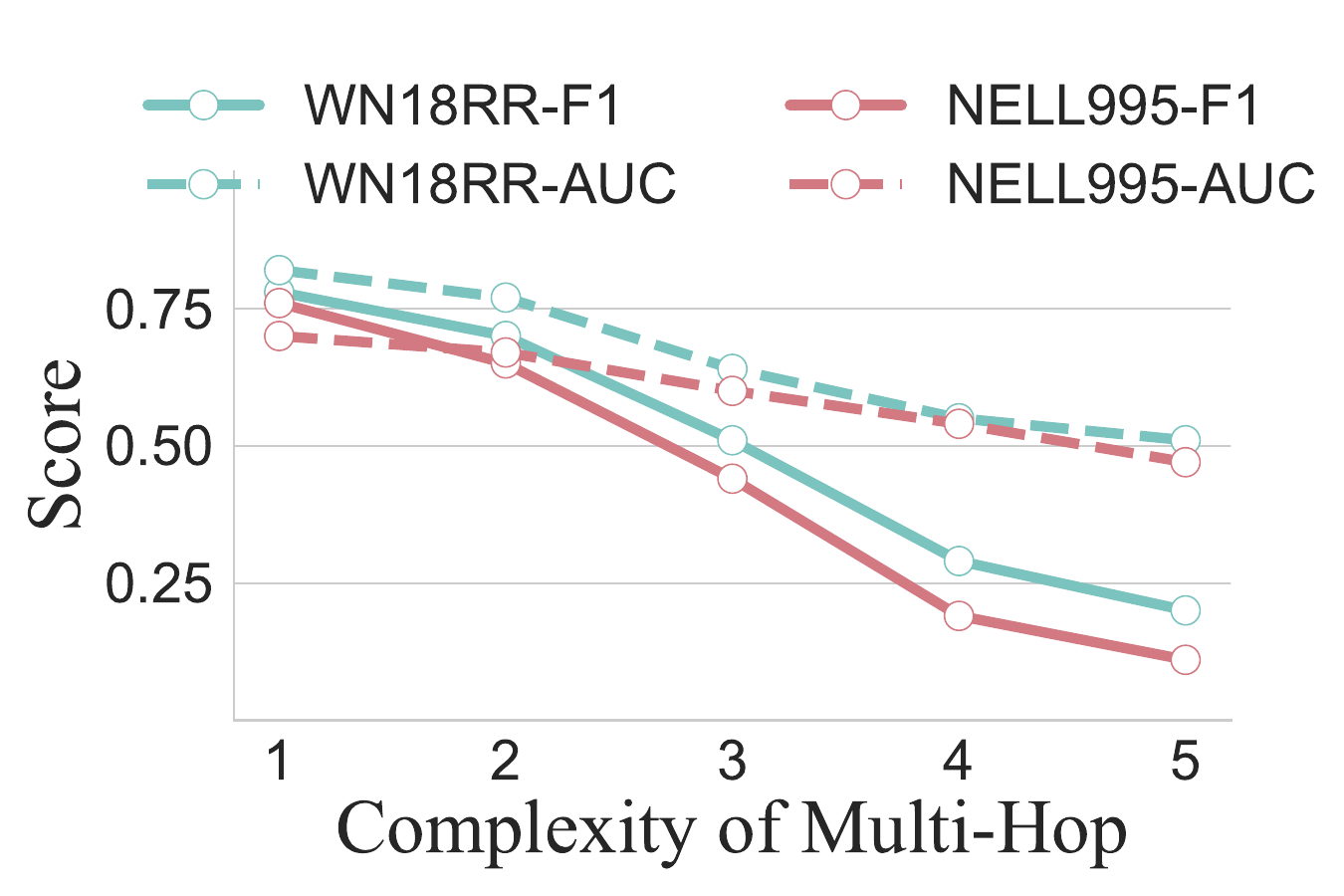}}
        \subfigure[ConGLR]{\includegraphics[width=0.32\linewidth]{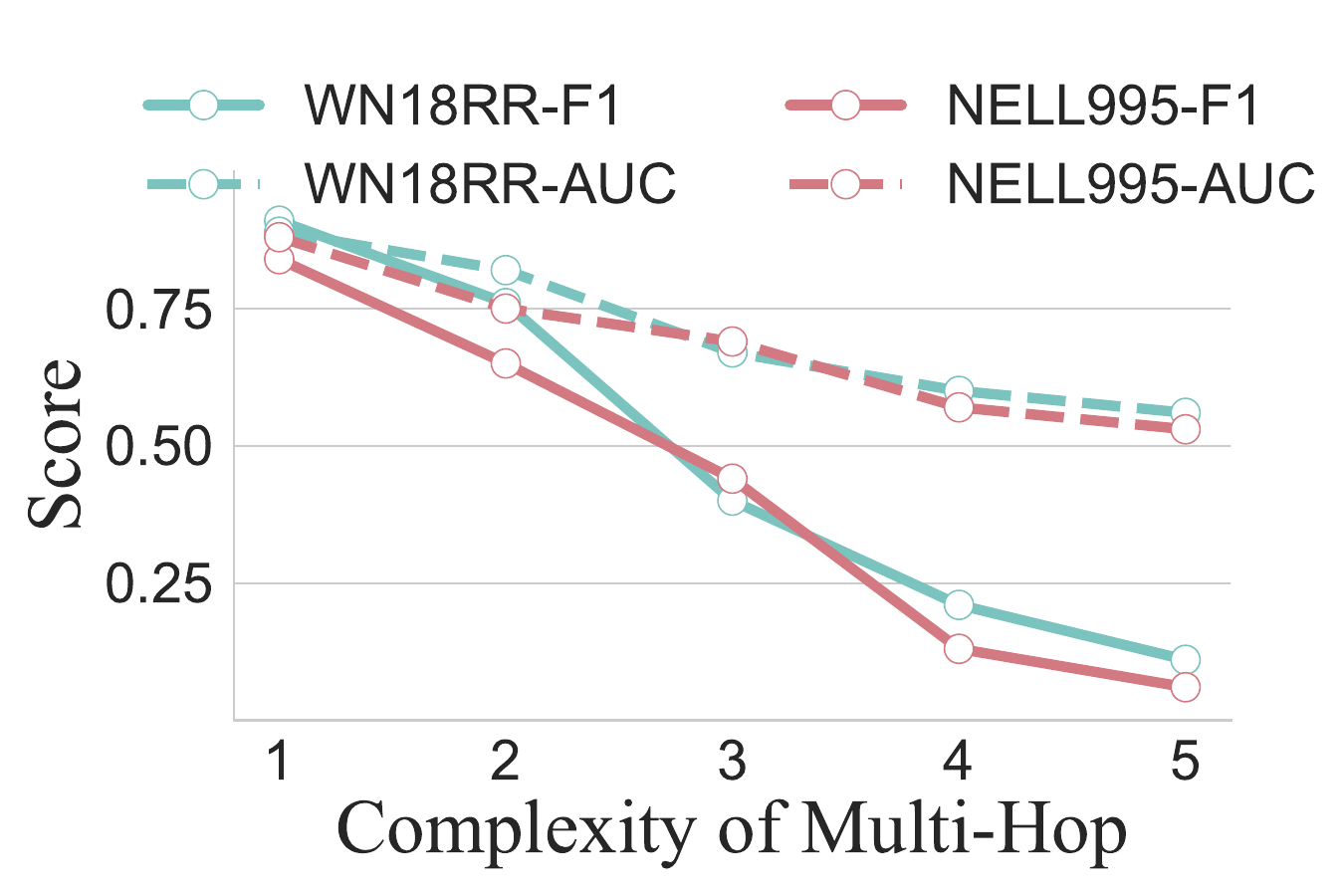}}
        \subfigure[ConvRot]{\includegraphics[width=0.32\linewidth]{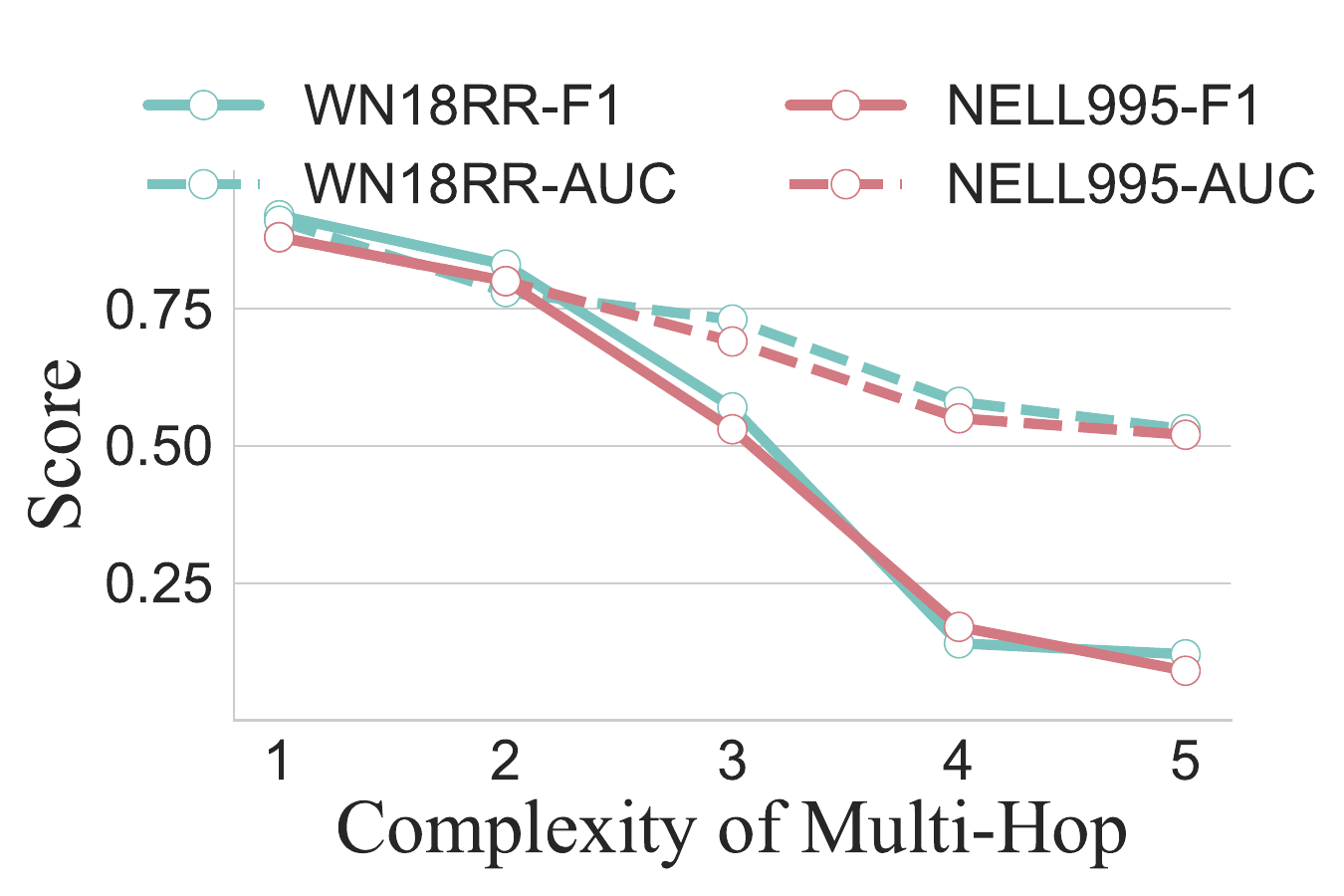}}
        
	\subfigure[Flan-T5 (Ablation)]{\includegraphics[width=0.32\linewidth]{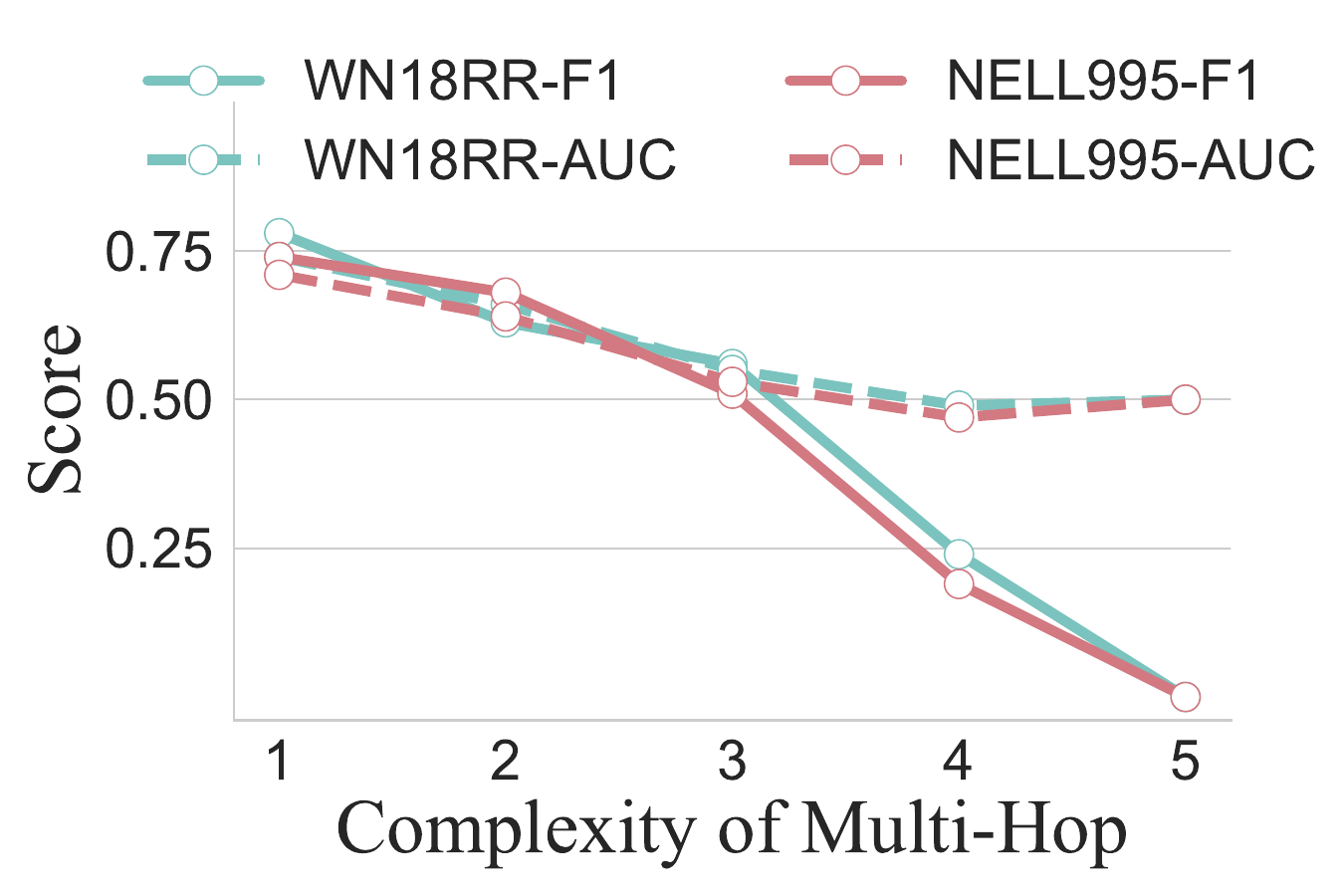}}
	\subfigure[LLaMa2 (Ablation)]{\includegraphics[width=0.32\linewidth]{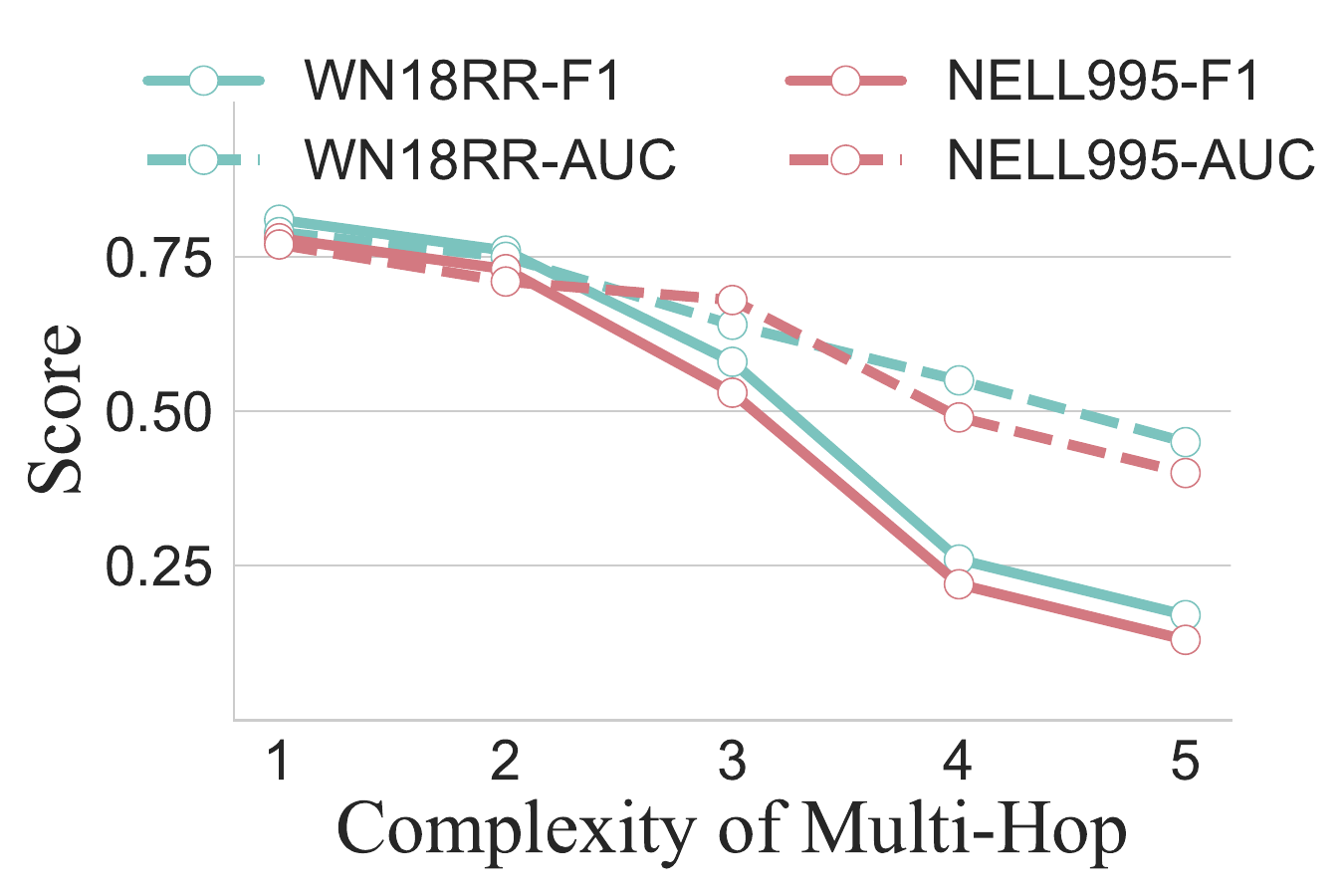}}
	\subfigure[Gemma (Ablation)]{\includegraphics[width=0.32\linewidth]{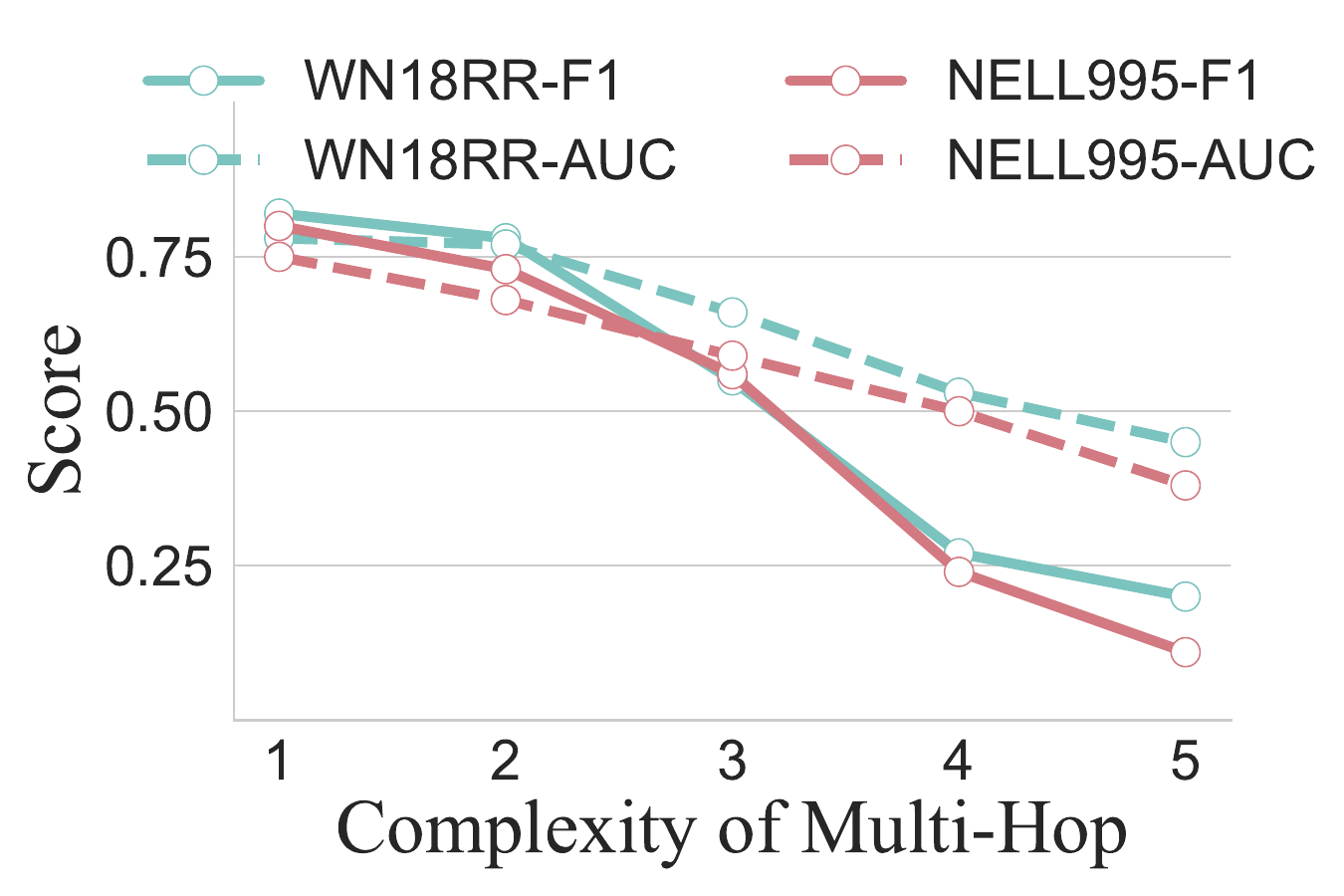}}
 
        \subfigure[Flan-T5 (KGLLM)]{\includegraphics[width=0.32\linewidth]{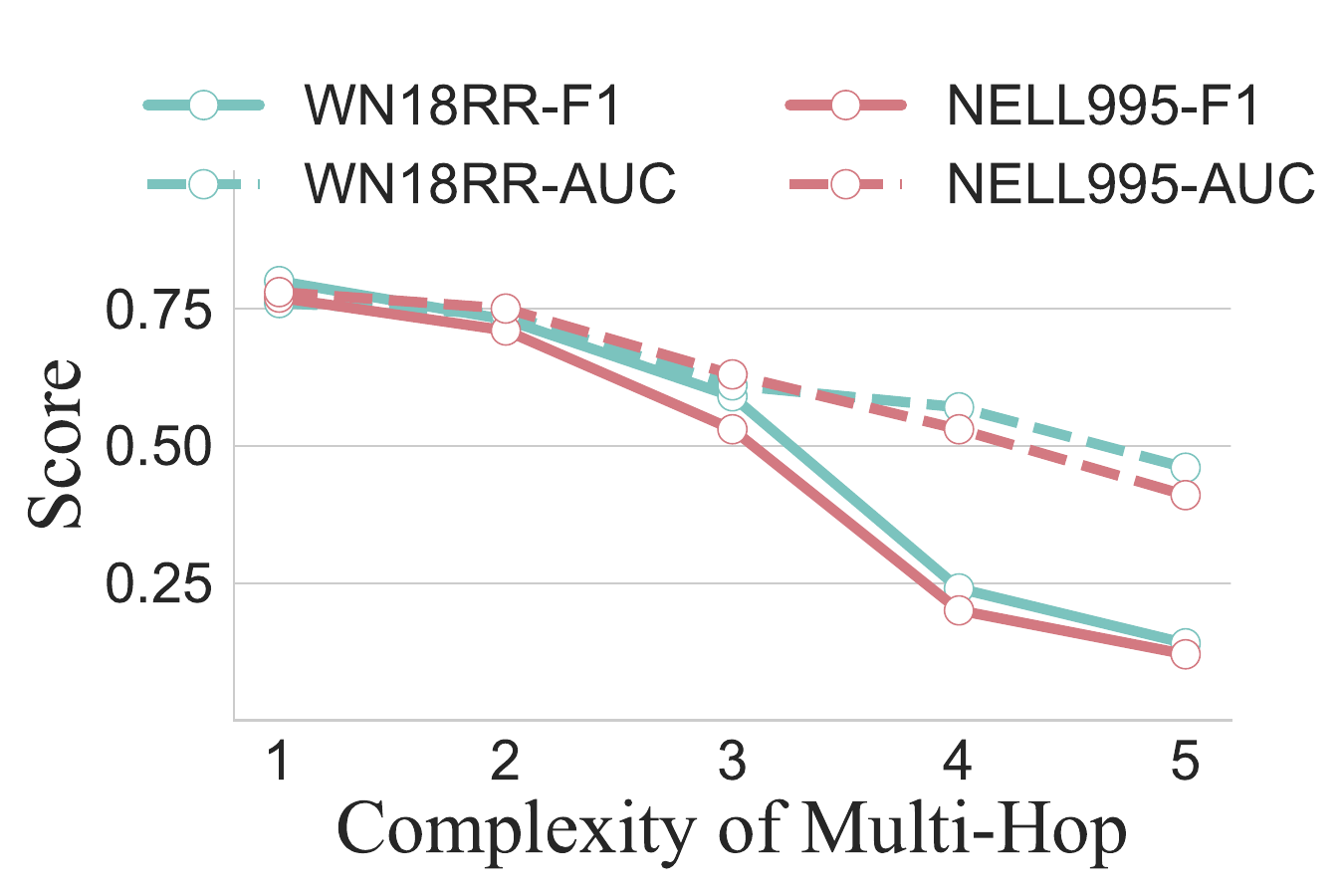}}
	\subfigure[LLaMa2 (KGLLM)]{\includegraphics[width=0.32\linewidth]{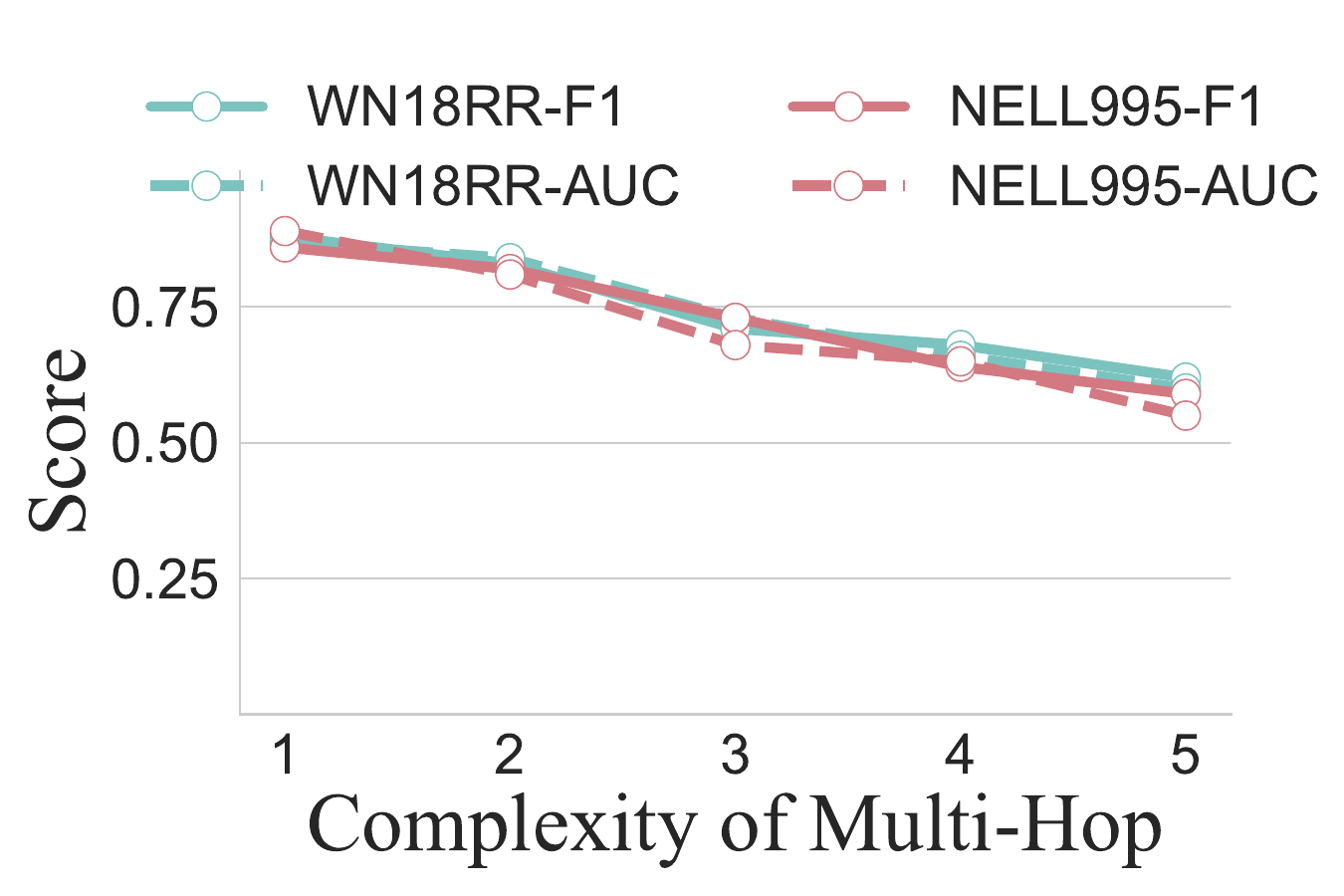}}
	\subfigure[Gemma (KGLLM)]{\includegraphics[width=0.32\linewidth]{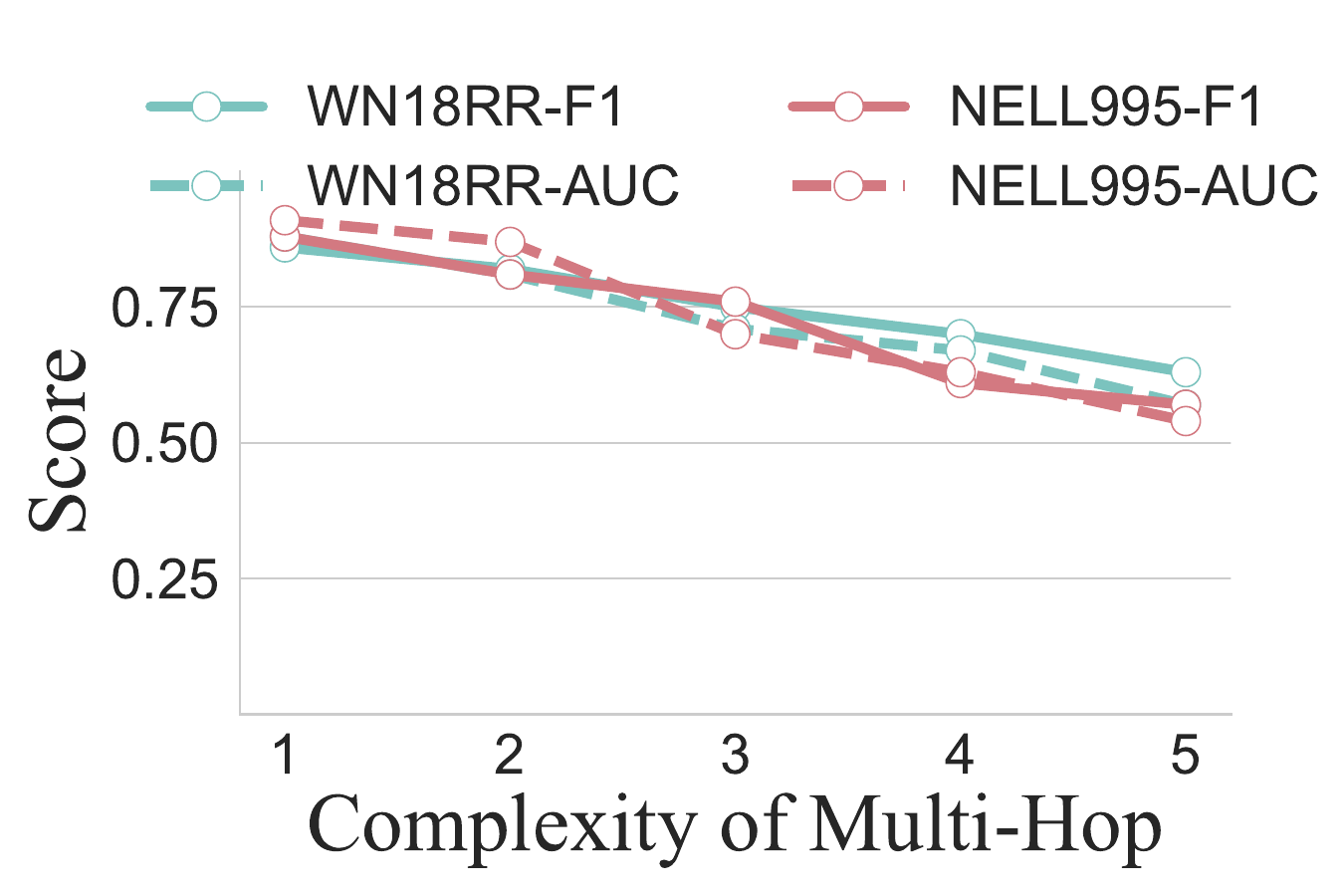}}
\caption{\textbf{Linear Relationship Between Complexity of Multi-Hop and Performance Score}}
\label{linear_relation}
\end{figure*}


\begin{table}[ht!]
\centering
\caption{Multi-hop Link Prediction \textbf{with} In-Context Learning}
\setlength{\tabcolsep}{2.5pt}
\begin{tabular}{ccccccccc}
\toprule
Datasets & \multicolumn{2}{c}{\textbf{WN18RR}} & \multicolumn{2}{c}{\textbf{NELL-995}} & \multicolumn{2}{c}{\textbf{FB15k-237}} & \multicolumn{2}{c}{\textbf{YAGO3-10}}\\ 
\cmidrule(lr){1-1}\cmidrule(lr){2-3}\cmidrule(lr){4-5}\cmidrule(lr){6-7}\cmidrule(lr){8-9}
Metrics & F1  & AUC  & F1  & AUC & F1  & AUC & F1  & AUC \\
\cmidrule{1-9}
Flan-T5 (Ablation) &0.70 &0.74 &0.61 &0.63& 0.55 & 0.63 & 0.57 & 0.63\\
LLaMa2 (Ablation) &0.83 &0.85 &0.81 &0.80& 0.76 & 0.78 & 0.84 & 0.83\\
Gemma (Ablation) &0.86 &0.87 &0.82 &0.83& 0.75 & 0.80 & 0.88 & 0.87\\
\cmidrule{1-9}
Flan-T5 (KG-LLM) & 0.85 & 0.87 &0.68 &0.70& 0.74 & 0.77 & 0.69 & 0.73\\
LLaMa2 (KG-LLM) & \underline{0.97} & \bf0.95 &\bf0.96 &\underline{0.93}& \underline{0.85} & \underline{0.86} & \bf0.95 & \underline{0.93}\\
Gemma (KG-LLM) & \bf 0.98 & \underline{0.94} &\underline{0.95} &\bf0.95& \bf0.94 & \bf0.92 & \underline{0.91} & \bf0.94\\

\bottomrule
\end{tabular}
\label{link_with}
\end{table}

\subsection{Multi-hop Link Prediction without In-Context Learning}
This section analyzes the traditional approaches, ablation framework, and KG-LLM framework in the context of non-In-Context Learning (non-ICL) Link Prediction, as shown in Table \ref{link_without}. Traditional approaches are shown in the top section of the table, the ablation framework is in the middle section, and the KG-LLM framework is in the bottom section.

\noindent
\textbf{Answer to Q1:} Our analysis reveals that the traditional approach's GNN model, especially ConvRot, exhibited relatively good performance, particularly surpassing the ablation models on the WN18RR dataset. This GNN model performance can be attributed to its ability to effectively capture the structural information in graph data. However, the results demonstrate that across all models, the implementation of the KG-LLM framework surpasses the traditional approaches and ablation framework across all datasets. This improved performance can be attributed to the KG-LLM framework's knowledge prompts. These prompts enable LLMs to take advantage of the relationships network between entities and their interconnections within the KG. Furthermore, these LLMs already possess basic common sense knowledge from pre-training. When all nodes and relations are converted to text, this inherent common sense enhances their understanding of the relations and nodes, thereby improving link prediction accuracy. Instruction fine-tuning (IFT) also contributed to this improvement by forcing models to focus on the limited options. The evidence presented here underscores the efficacy of our KG-LLM framework, enriched with CoT and IFT, indicating its potential to advance the domain of multi-hop link prediction tasks in real-world applications.

We also evaluate GNN, ablation, and KG-LLM framework models' performance at each level of hop complexity in WN18RR and NELL-995 datasets. As shown in Figure \ref{linear_relation}, the performance of GNN and ablation models significantly declines as hop complexity increases. Upon closer examination, it is evident that as hop complexity grows, these models frequently respond with 'No' for most questions, resulting in an F1 score close to 0 and an AUC score around 0.5. This performance is due to the increased complexities of multi-hop link prediction. Unlike the straightforward task of predicting a direct link between two nodes, models must consider all intermediate nodes to conclude, adding significant complexity and reducing their effectiveness. In contrast, the KG-LLM framework models effectively address this challenge, maintaining fair performance even at five-hops, except for the Flan-T5 model.

\subsection{Multi-hop Link Prediction with In-Context Learning}

In this section, we evaluate the influence of In-Context Learning (ICL) on models subjected to both ablation and KG-LLM frameworks, excluding the traditional approach as it lacks ICL capability. We experimented using the same LLMs and testing inputs as in the previous section. The key distinction in this evaluation was adding an ICL example at the beginning of each original testing input. The ICL example shown in Appendix \ref{appendix:icl}, derived from the training dataset, was restricted to the complexity of two-hops. This constraint aimed to prevent providing additional knowledge through the ICL example while furnishing a contextually relevant example.

Table \ref{link_with} reveals a notable enhancement in the performance of models under the ablation framework, with LLaMa2 and Gemma models achieving an F1 and AUC score exceeding 80\% in WN18RR and NELL-995 datasets. Remarkably, the adoption of ICL within the KG-LLM framework resulted in a significant performance uplift. Notably, the Gemma model achieved a staggering 98\% F1 score on the first dataset, while LlaMa2 recorded a 96\% F1 score on the second dataset.

An interesting observation is that ICL has shown unstable improvements in the Flan-T5 model. For some datasets within the ablation and KG-LLM frameworks, the performance slightly declined after implementing ICL. This phenomenon could be attributed to the increased length and complexity of the testing prompts. While the inclusion of an ICL example generally aids in model understanding, in certain cases, it might be perceived as noise, potentially affecting the Flan-T5's performance.

\noindent
\textbf{Answer to Q2:} The experimental results indicate that the deployment of ICL does not uniformly improve performance across all models. However, for the LLaMa2 and Gemma models, the integration of ICL consistently facilitates performance improvements.

\begin{figure*}[!h]
	\centering  
        \captionsetup[subfigure]{skip=2pt}
        
        \subfigure[Ablation Framework]{\includegraphics[width=0.49\linewidth]{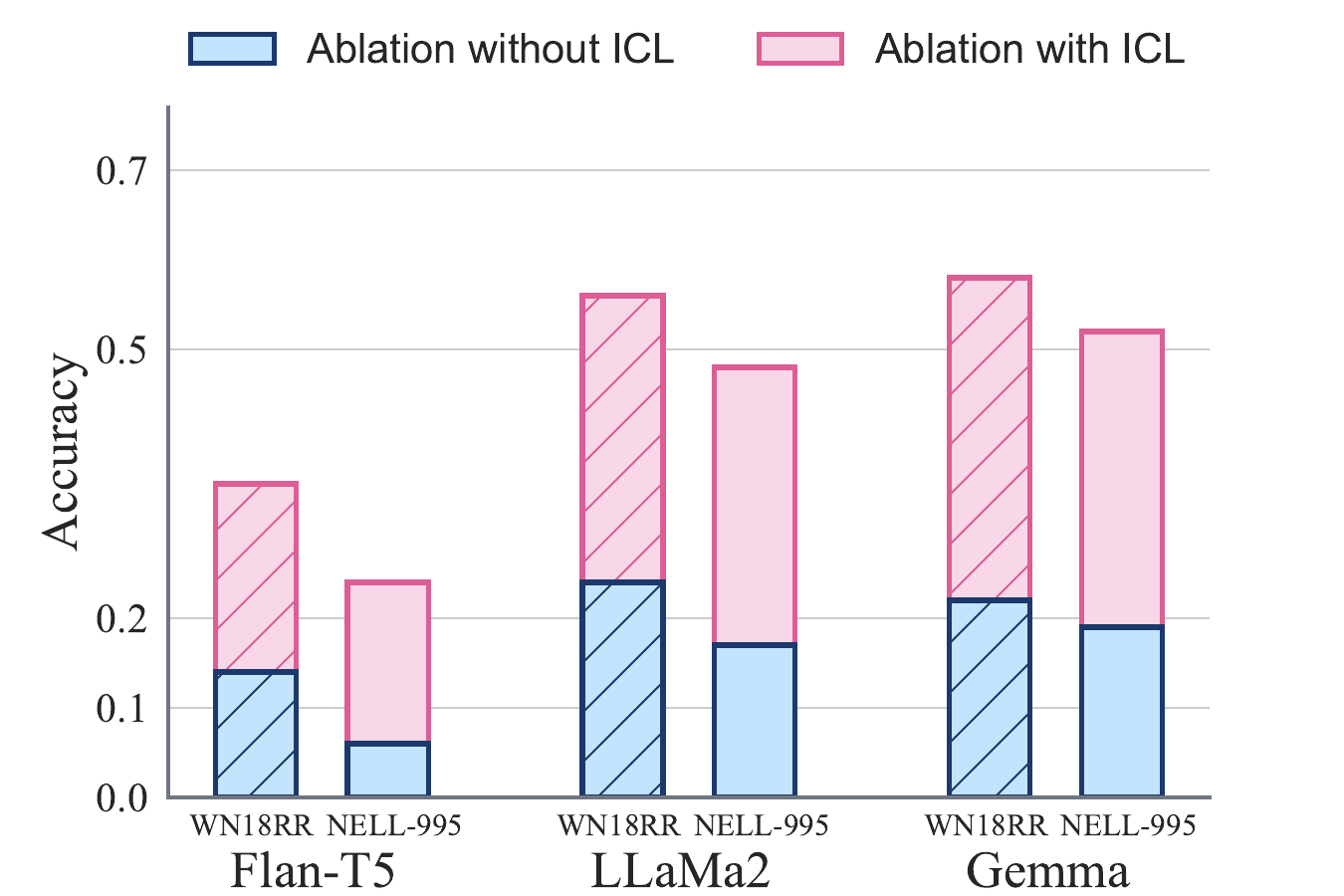}}
        \subfigure[KGLLM Framework]{\includegraphics[width=0.49\linewidth]{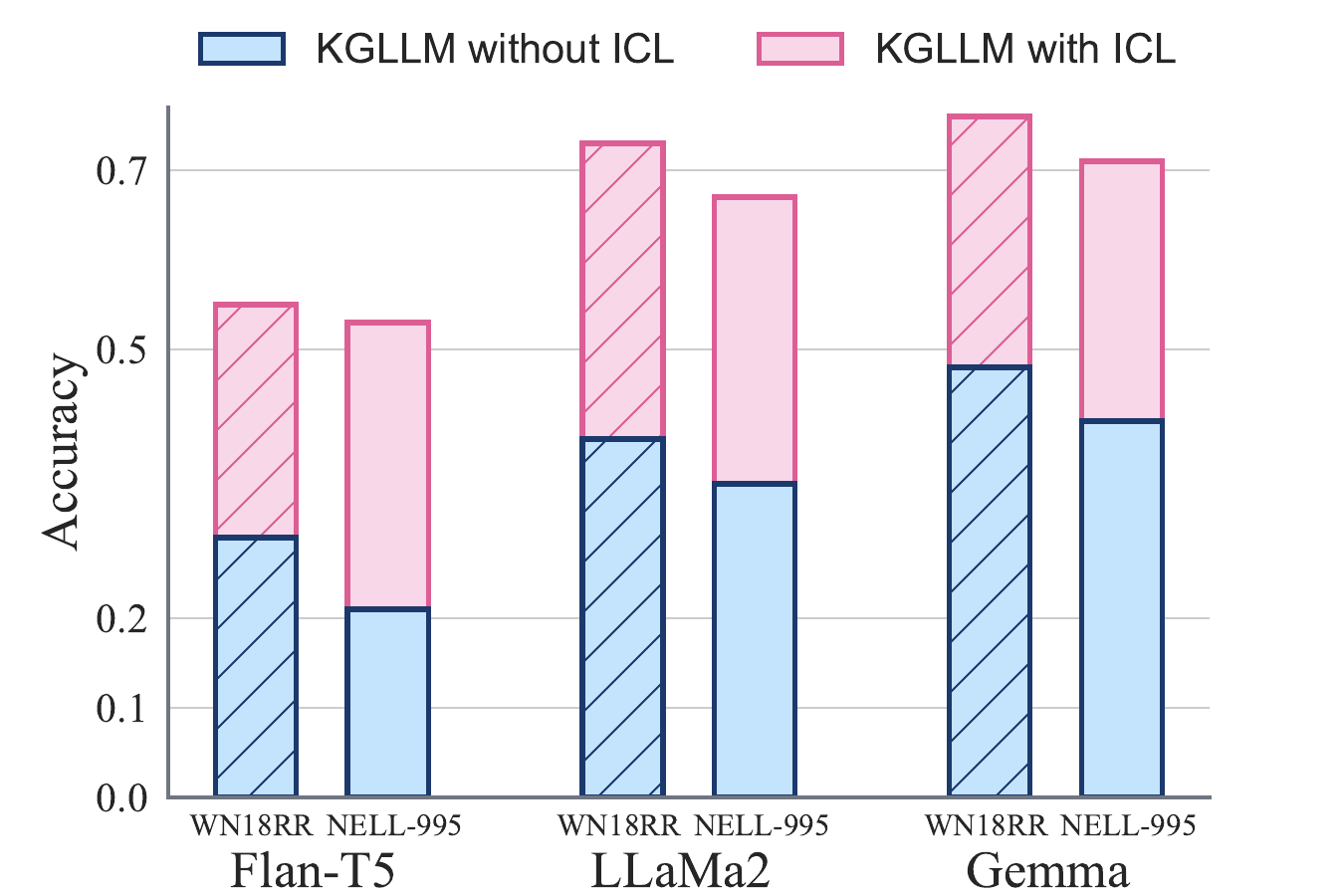}}

\caption{\textbf{Multi-hop Relation Prediction Performance Comparison:} The left graph shows model performance under the ablation framework, while the right graph shows model performance under the KGLLM framework. Blue bars represent testing without ICL, and red bars represent testing with ICL.}
\label{relation_prediction}
\end{figure*}

\subsection{Multi-hop Relation Prediction without In-Context Learning}

In this analysis, we explore models' ability to perform unseen multi-hop relation prediction tasks on WN18RR and NELL-995 datasets, excluding the traditional approach as it lacks generalization ability. We used the same testing dataset in the multi-hop link prediction task to ensure comparability and fairness. As mentioned in section \ref{knowledge_prompt}, the difference lies in the instruction and prompt question presented to the model. 

Our findings are presented in Table \ref{relation_prediction}. We discovered that both frameworks showcased limited performance in this task without ICL. Notably, the KG-LLM framework exhibited marginally superior performance. Upon reviewing the predictive results, we observed that the model continues to provide `yes' and `no' answers for most questions, similar to the multi-hop link prediction task. For some questions, it outputs random responses.

\noindent
\textbf{Answer to Q3:} The findings suggest that the KG-LLM framework marginally enhances the models' generalization abilities. However, it would be premature to assert that our framework equips models with the ability to navigate unseen prompts. This could be attributed to the complexity and difficulty of the new instructions and options. With options no longer limited to a binary yes or no answer, the model may struggle to comprehend the updated instruction and effectively utilize the provided options.

\subsection{Multi-hop Relation Prediction with In-Context Learning}

We further explore the impact of incorporating ICL into the multi-hop relation prediction task. The ICL example is shown in the Appendix \ref{appendix:icl}. The results of red bars (with ICL) in Table \ref{relation_prediction} reveal a significant improvement in the generalization abilities of the models under both ablation and KG-LLM frameworks, in contrast to the blue bars (without ICL). In particular, the LlaMa2 and Gemma models under the KG-LLM framework with ICL, achieved an accuracy exceeding 70\% in the WN18RR datasets.

\noindent
\textbf{Answer to Q4:} The integration of ICL has improved the models' ability to excel in unseen tasks. The KG-LLM framework, in particular, exhibits the ability to learn and utilize the contextual example provided by ICL.

\section{Conclusions and Future Work}

Our study introduces the Knowledge Graph Large Language Model Framework (KG-LLM Framework) as a promising solution for multi-hop generative prediction tasks in knowledge graph analysis. By leveraging techniques such as Chain of Thought and Instruction Fine-tuning, models processed by KG-LLM framework have greatly enhanced the accuracy of predictions in KG-related tasks. 


In our future work, we will consider accessing the model's reasoning process in the evaluation stage. We are committed to further enhancing KG-related prediction tasks. One key aspect we will focus on is refining the instruction process by limiting the option size. Additionally, we plan to explore the utilization of prompt quality filters to effectively filter out noisy options, improving the overall accuracy and reliability of the model's predictions. Through these ongoing improvements, we aim to advance the capabilities of KG-LLM models and contribute to the progress of knowledge graph analysis.



\bibliography{acml24}

\newpage

\appendix

\section{Appendix}\label{apd:first}


\subsection{In-Context Learning Examples}
\label{appendix:icl}
\subsubsection{Multi-hop Link Prediction ICL Example in Ablation Framework}

\#\#\# Context:\newline
Node\_47405 has relation\_179 with Node\_46497. Node\_46497 has relation\_180 with Node\_46501. Is Node\_47405 connected with Node\_46501?\newline
Answer:\newline
The answer is yes.

\subsubsection{Multi-hop Link Prediction ICL Example in KG-LLM Framework}

\#\#\# Context:\newline
Node\_47405 has relation\_179 with Node\_46497. Node\_46497 has relation\_180 with Node\_46501. Is Node\_47405 connected with Node\_46501?\newline
Answer:\newline
Node\_47405 has relation\_179 with Node\_46497, means Miles Davis music artist is associated with genre Bebop. Node\_46497 has relation\_180 with Node\_46501, means Bebop genre is under the broader genre Jazz. So Miles Davis music artist is associated with genre Jazz.\newline
The answer is yes.

\subsubsection{Multi-hop Relation Prediction ICL Example in Ablation Framework}

\#\#\# Context:\newline
Node\_47405 has relation\_179 with Node\_46497. Node\_46497 has relation\_180 with Node\_46501. What is the relationship between Node\_47405 and Node\_46501?\newline
Answer:\newline
The relationship between the Node\_47405 and Node\_46501 is relation\_179.

\subsubsection{Multi-hop Relation Prediction ICL Example in KG-LLM Framework}

\#\#\# Context:\newline
Node\_47405 has relation\_179 with Node\_46497. Node\_46497 has relation\_180 with Node\_46501. What is the relationship between Node\_47405 and Node\_46501?\newline
Answer:\newline
Node\_47405 has relation\_179 with Node\_46497, means Miles Davis music artist is associated with genre Bebop. Node\_46497 has relation\_180 with Node\_46501, means Bebop genre is under the broader genre Jazz. So Miles Davis music artist is associated with genre Jazz.\newline
The relationship between Miles Davis and Jazz is music\_artist\_genre. 

\newpage
\subsection{Dataset 2}

\subsubsection{ablation Framework One-shot In-Context Learning for Multi-hop Link Prediction}
\ 
\#\#\# Context:\newline
node\_11809 has relation\_9 with node\_12218. node\_12218 has relation\_1 with node\_9431. Is node 11809 connnected with node 9431?\newline
Answer:\newline
The answer is yes.

\subsubsection{KG-LLM Framework One-shot In-Context Learning for Multi-hop Link Prediction}
\ 
\#\#\# Context:\newline
node\_11809 has relation\_9 with node\_12218. node\_12218 has relation\_1 with node\_9431. Is node 11809 connnected with node 9431?\newline
Answer:\newline
node\_11809 has relation\_9 with node\_12218, means node\_11809 \_verb\_group node\_12218. node\_12218 has relation\_1 with node\_9431, means node\_12218 \_derivationally\_related\_form node\_9431. So node 11809 \_derivationally\_related\_form node 9431. The answer is yes.

\subsubsection{ablation Framework One-shot In-Context Learning for Multi-hop Relation Prediction}
\ 
\#\#\# Context:\newline
node\_11809 has relation\_9 with node\_12218. node\_12218 has relation\_1 with node\_9431. What is the relationship between node\_11809 and node\_9431?\newline
Answer:\newline
The relationship between the first node and the last node is relation\_1.

\subsubsection{KG-LLM Framework One-shot In-Context Learning for Multi-hop Relation Prediction}
\ 
\#\#\# Context:\newline
node\_11809 has relation\_9 with node\_12218. node\_12218 has relation\_1 with node\_9431. What is the relationship between node\_11809 and node\_9431?\newline
Answer:\newline
node\_11809 has relation\_9 with node\_12218, means node\_11809 \_verb\_group node\_12218. node\_12218 has relation\_1 with node\_9431, means node\_12218 \_derivationally\_related\_form node\_9431. So node 11809 \_derivationally\_related\_form node 9431. The relationship between the first node and the last node is \_derivationally\_related\_form.


 

\end{document}